\documentclass[a4paper,fleqn]{cas-dc}

\usepackage[numbers]{natbib}
\usepackage{subfigure}

\def\tsc#1{\csdef{#1}{\textsc{\lowercase{#1}}\xspace}}
\tsc{WGM}
\tsc{QE}
\tsc{EP}
\tsc{PMS}
\tsc{BEC}
\tsc{DE}

\begin{document}
\let\WriteBookmarks\relax
\def\floatpagepagefraction{1}
\def\textpagefraction{.001}
\shorttitle{Purin: A Biology-Inspired Mechanism for ANNs}
\shortauthors{Zishu Liu et~al.}

\title [mode = title]{Purin: a biology-inspired mechanism for artificial neural networks}                      



\author[1]{Zishu Liu}[style=chinese,
                        orcid=0009-0009-4179-0508]
\ead{zl538@exeter.ac.uk}

\credit{Conceptualization of this study, Methodology, Software}

\affiliation[1]{organization={Department of Computer Science, University of Exeter},
                city={Exeter},
                postcode={EX4 4QJ}, 
                state={Devon},
                country={United Kingdom}}

\author[1]{Chunbo Luo}[style=chinese]
\ead{C.Luo@exeter.ac.uk}
\cormark[1]

\author[2]{Christos Grecos}
\ead{grecoschristos@gmail.com}



\affiliation[2]{organization={Department of Computer Science, University of Wisconsin Parkside},
                city={Kenosha},
                postcode={53144},
                State={Wisconsin,},
                country={USA}
                }



\cortext[cor1]{Corresponding author}


\begin{abstract}
Artificial neural networks (ANNs) usually represent neural transmission with fixed trainable weights during a training batch, which omits short-term changes in synaptic efficacy. In addition, the discrete time-step simulation requires additional temporal processing that many conventional ANN architectures do not use. To overcome these challenges, we propose Purin, a biology-inspired and ANN-compatible mechanism, that introduces synaptic efficacy modulation into conventional convolutional neural networks. Purin uses a time-interval-based abstraction for neural activities, which allows Purin to introduce short- and long-term synaptic efficacy changes without using discrete time-steps. Purin introduces a bounded short-term factor interpreted as temporary synaptic efficacy changes, together with two trainable components that represent input-side and output-side efficacy, respectively. These components are updated by backpropagation and interpreted as the long-term synaptic efficacy changes. Experimental results show that after removing the confounding factors in the AlexNet, VGG11, and GoogLeNet architectures, Purin improves the classification accuracies in all three models in the evaluated datasets with a minor increase in model size and computational complexity. The controlled input order experiments show that the short-term factor generates history-dependent responses to the same input images.
\end{abstract}




\maketitle

\section{Introduction}

Artificial neural networks (ANNs) have been widely used in many areas, such as image classification \cite{AlexNet} and segmentation \cite{WANG2020135}, natural language processing \cite{NIPS2017_3f5ee243}, and object detection \cite{KAUR2023103812}. Although they have achieved promising performance in many areas, many ANNs are biologically implausible in various aspects such as the structure of neurons, the activation rule (action potential vs. activation functions), the learning rule, etc. Researchers try to narrow those gaps by proposing neural networks that introduce biological mechanisms. One of the most popular families is the spiking neural network (SNN) family, which introduces biologically plausible spiking neurons. However, they face difficulties in training the network because the spike generation functions are not differentiable. Thus, SNNs cannot directly use the backpropagation (BP) algorithm \cite{tavanaei2019deep}.

In this paper, we propose a biology-inspired mechanism called Purin to reduce selected gaps between ANN and biological neurons while making it compatible with representative ANN architectures and the BP learning rule. Firstly, we develop an abstraction method for biological neurons, which introduces neurons with two weight components that reflect the efficacy of the input-side and output-side efficacy terms, respectively. Secondly, Purin introduces an abstract concept of time intervals by treating a still image as information presented in a short time interval. It allows Purin to introduce short- and long-term weight dynamics without extra transformation steps (e.g., spike encoding). Thirdly, inspired by short- and long-term plasticity mechanism in biology, we introduce a short-term factor $g_{stp}$ in Purin that introduces activation-dependent changes to the network. The short-term factor causes temporary weight changes within a batch, which also affects the long-term update based on BP in the weight components.

The rest of the paper is organized as follows: Section \ref{related} briefly introduces existing research on biology-inspired deep learning models, dynamic synapses, and gating and stochastic modulation in deep learning. Section \ref{method} introduces the proposed method in detail. Section \ref{experiments} shows the results related to $g_{stp}$ behaviours, compatibility experiments, and ablation studies. Conclusions and future work are given in Section \ref{conclusion}.

\section{Related work}
\label{related}
\subsection{Biology-inspired Neural Network}
ANNs are inspired by biological neurons, but focus more on principles such as function optimization than biology \cite{marblestone2016toward}. Despite the fact that researchers tried to explain or approximate the BP algorithm in various ways \cite{marblestone2016toward, whittington2019theories}, the differences between biological neurons and ANN neurons, such as spike-based versus continuous-vector outputs and distinct learning rules, eventually motivate the development of the SNN family.

SNNs introduce a more biologically plausible neuron model that emits spikes when the membrane potential reaches the firing threshold. The synapses (or weights) in SNNs are updated based on biological principles like the Hebbian rule \cite{hebb2005organization} and spike-timing dependent plasticity \cite{stdp} rather than the BP algorithm. However, training deep SNNs based on discrete spikes is challenging because it cannot use derivative-based optimization methods like ANNs \cite{taherkhani2020review}. To overcome this difficulty, researchers proposed the surrogate gradient for SNNs training \cite{zenke2018superspike, bellec2018long}. Another method to address the training difficulty in deep SNN is ANN-to-SNN conversion, which converts pre-trained ANN models to SNN models to make SNN models compatible with existing ANN models and reduce the resources for training \cite{hu2024toward}. However, deep SNNs still face the gradient vanishing issue in the deep layers due to the binary spikes \cite{hu2024toward}. In addition to the training difficulty, SNNs require encoding images into spikes, which is less common for many existing image classification datasets. As a result, SNNs usually need additional preprocessing steps in comparison to ANNs. Although researches have tried the idea of combining a CNN feature extractor and an SNN to make final decisions \cite{xu2022hierarchical}, the additional spike encoding is still necessary in this model.

\subsection{Short-term Plasticity}
The short-term plasticity (STP) of synapses has been explored by many researchers. In \cite{abbott1997synaptic}, the authors suggested that the short-term depression of synapses provides a dynamic gain-control mechanism. The short-term depression can be described by a multiplicative factor in the model mentioned in that paper. In \cite{tsodyks1998neural}, the authors developed mathematical models for the computation of short-term synaptic plasticity. A survey \cite{hennig2013theoretical} summarizes theoretical models of short-term synaptic plasticity. These models focus on different points that cause the short-term synaptic plasticity, and many of them suggest that this mechanism can be written as a multiplicative factor in mathematical equations. However, the models mentioned in \cite{tsodyks1998neural, hennig2013theoretical} are usually based on spiking or discrete time steps, which is incompatible with the requirement of continuous inputs or outputs in ANNs.

Researchers tried to introduce STP to ANNs without using spiking inputs or outputs. In STPN \cite{STPN}, the authors introduced an additive component to a neuron's weights, which played the role of STP in synaptic efficacy. However, STPN is designed for the RNN architecture, which has a different neuron structure from ANN neurons. Although hardware implementations of the STP mechanism have been reported in \cite{weilenmann2024single}, this paper focuses on software-based implementations through algorithm design. 

\subsection{Gating Mechanism in ANN}
In deep learning, attention and gating methods are usually used to learn sample-dependent features. The Squeeze-and-Excitation (SE) block proposed in \cite{SEblock} performs channel-wise feature recalibration by modeling interdependencies between the convolution channels. However, the SE block neglects important position information in the attention maps. Thus, coordinate attention \cite{hou2021coordinate} is proposed to capture both channel-wise information and long-range positional information. In terms of gating methods, the long short-term memory \cite{LSTM} introduced three gates (the input gate, the forget gate and the output gate) to enable the network to keep important long-term memory information. The gMLP \cite{gmlp} introduced a spatial gating unit, which used a linear projection matrix independent of the inputs of unit, to allow cross-token interactions. Despite this technique being successful, many feedforward attention and gating methods compute modulation values based on current inputs, but do not maintain an evolving state across successive samples. In contrast, $g_{stp}$ in Purin evolves according to preceding neural activity within a batch of samples.

\section{Method}
\label{method}
In this section, we describe the proposed Purin mechanism in detail. In general, an ANN neuron with the Purin mechanism is formulated as Eq (\ref{general}) where $X$ is its input, $W_{in}$ is the input-side weight matrix and $W_{out}$ is a vector representing the output-side efficacy of the neuron and $f(\cdot)$ is the signed magnitude function, which is a signed activity measure over a short time interval. The symbol $\odot$ denotes the Hadamard product. The variable $g_{stp}$ is a factor that reflects the temporary efficacy changes caused by the short-term plasticity. In fully connected layers, $g_{stp}$ is a vector that has the same shape as the layer's output vector. In 2D convolution layers, $g_{stp}$ is a vector with $C$ elements that matches the number of convolution kernels in that layer, which performs channel-wise multiplication in the convolution layers.
\begin{equation}
    \label{general}
    Y=g_{stp}\odot W_{out}\odot f(W_{in}X)
\end{equation}

In the following sections, we introduce our motivation, the abstraction of biological neurons, the short-term plasticity, and justify why there is no bias vector and dropout in our mechanism.

\subsection{Motivation}
\label{motivation}
The motivation of Purin is to narrow the selected gaps between ANNs and biological neurons, while preserving compatibility with continuous activations and the BP algorithm. Thus, Purin is not intended to be a biophysically faithful model of synaptic plasticity. Instead, Purin is designed based on two biological observations: the synaptic efficacy changes temporally due to the short-term plasticity, and the signal transmission between two neurons depends on the efficacy of pre- and post-synaptic components of the synaptic connection.

\begin{figure}
    \centering
    \includegraphics[width=\linewidth]{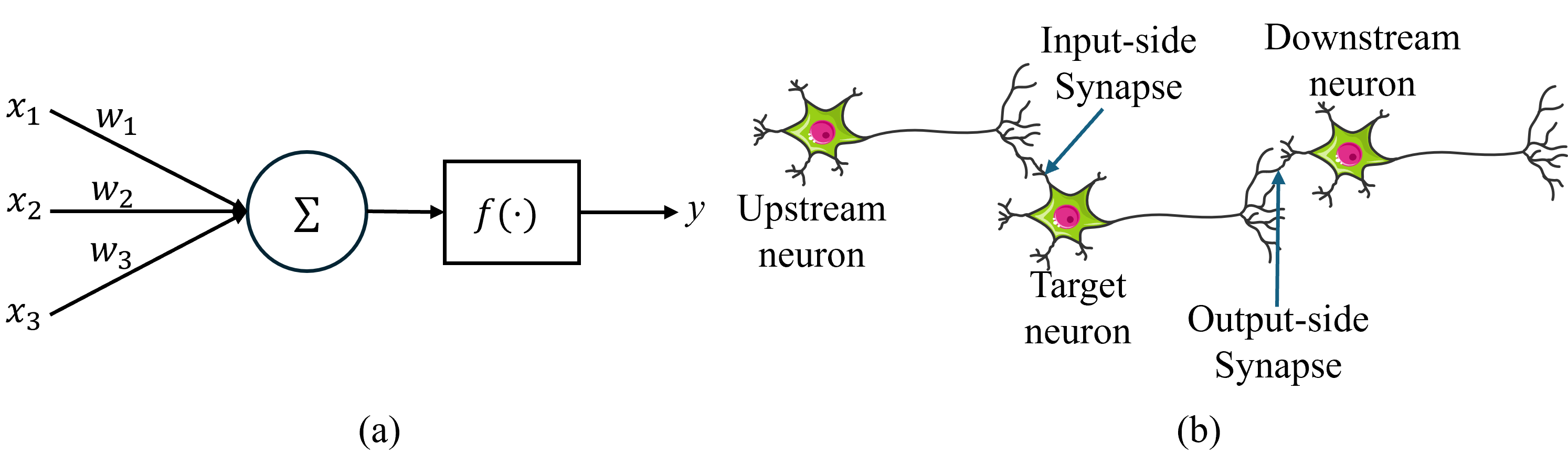}
    \caption{A schematic diagram of neurons in ANNs and biology. (a) A conventional neuron in ANN. (b) A biological target neuron that receives information from an upstream neuron through an input-side synapse and transmits information through an output-side synapse to a downstream neuron. Fig. 1 (b) is adapted from Servier Medical Art (https://smart.servier.com), licensed under CC BY 4.0 (https://creativecommons.org/licenses/by/4.0/)}
    \label{fig:difference}
\end{figure}

\begin{figure}
    \centering
    \includegraphics[width=\linewidth]{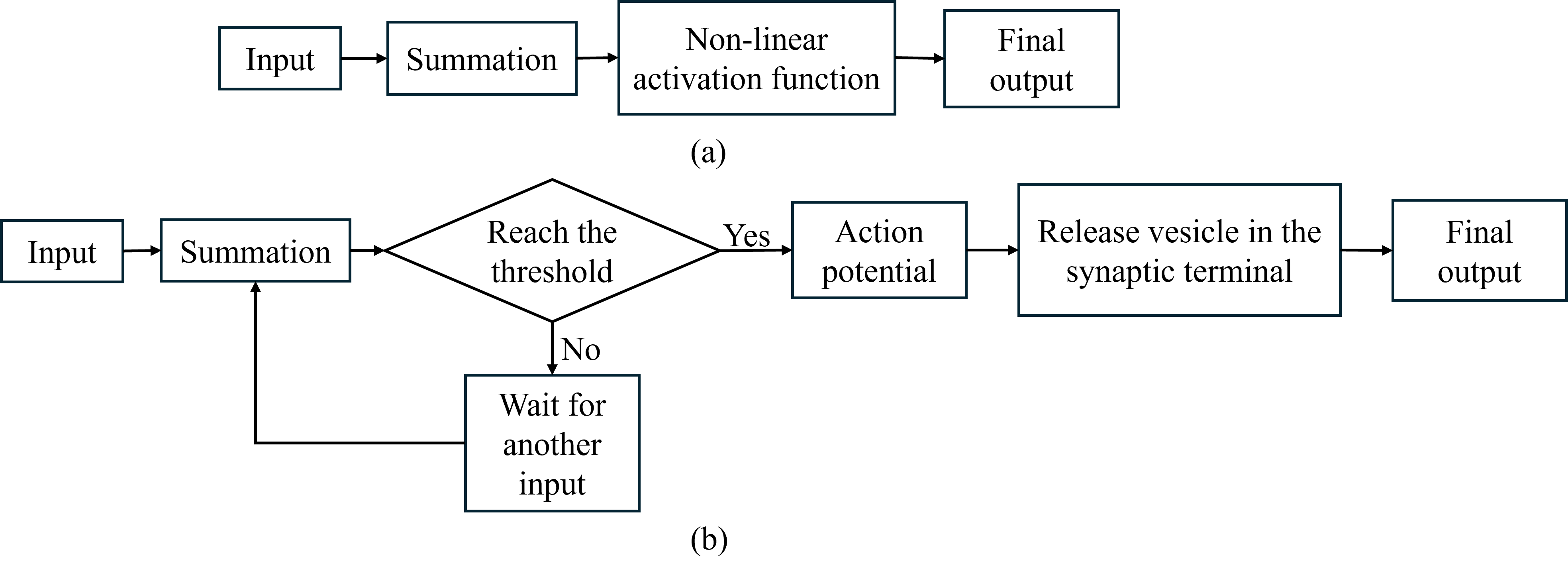}
    \caption{A flowchart describing how an ANN's neuron and biological neuron process their inputs. We focus on the biological neuron using chemical synapses. (a) ANN's neuron's procedure. (b) biological neuron's procedure. }
    \label{fig:diffproc}
\end{figure}

As shown in Fig. \ref{fig:difference}, neurons in ANNs usually do not have output weights after generating the output $y$. By contrast, a biological chemical synapse involves both presynaptic and postsynaptic components that contribute to transmission efficacy. This structural difference further leads to differences in transmitting the neuron's outputs (see Fig. \ref{fig:diffproc}). ANN neurons simplify the procedure of transmitting the action potential in biological neurons. However, the synaptic efficacy of input-side synapses plays a role in information transmission between biological neurons, and the synapse plasticity mechanism also works on the input-side synapses and output-side synapses separately. Thus, simplifying presynaptic and postsynaptic components into one ignores the enhancement or attenuation caused by the upstream neuron when transmitting the information. In addition, the trainable weights in ANNs are usually fixed during a training batch, while synapses in biological neurons show temporal dynamic changes due to the short-term plasticity mechanism, which can eventually lead to long-term changes in synaptic efficacy.

In Purin, we introduce a simplified form of synaptic efficacy modulation into CNN architectures without requiring spike generation, spike encoding, or discrete temporal simulation. The following subsections describe how output-side efficacy, abstract time intervals, and short-term modulation are formulated in Purin.

\subsection{Biology-inspired Abstraction of Neuron}
\label{bioabstract}
\begin{figure}
    \centering
    \includegraphics[width=\linewidth]{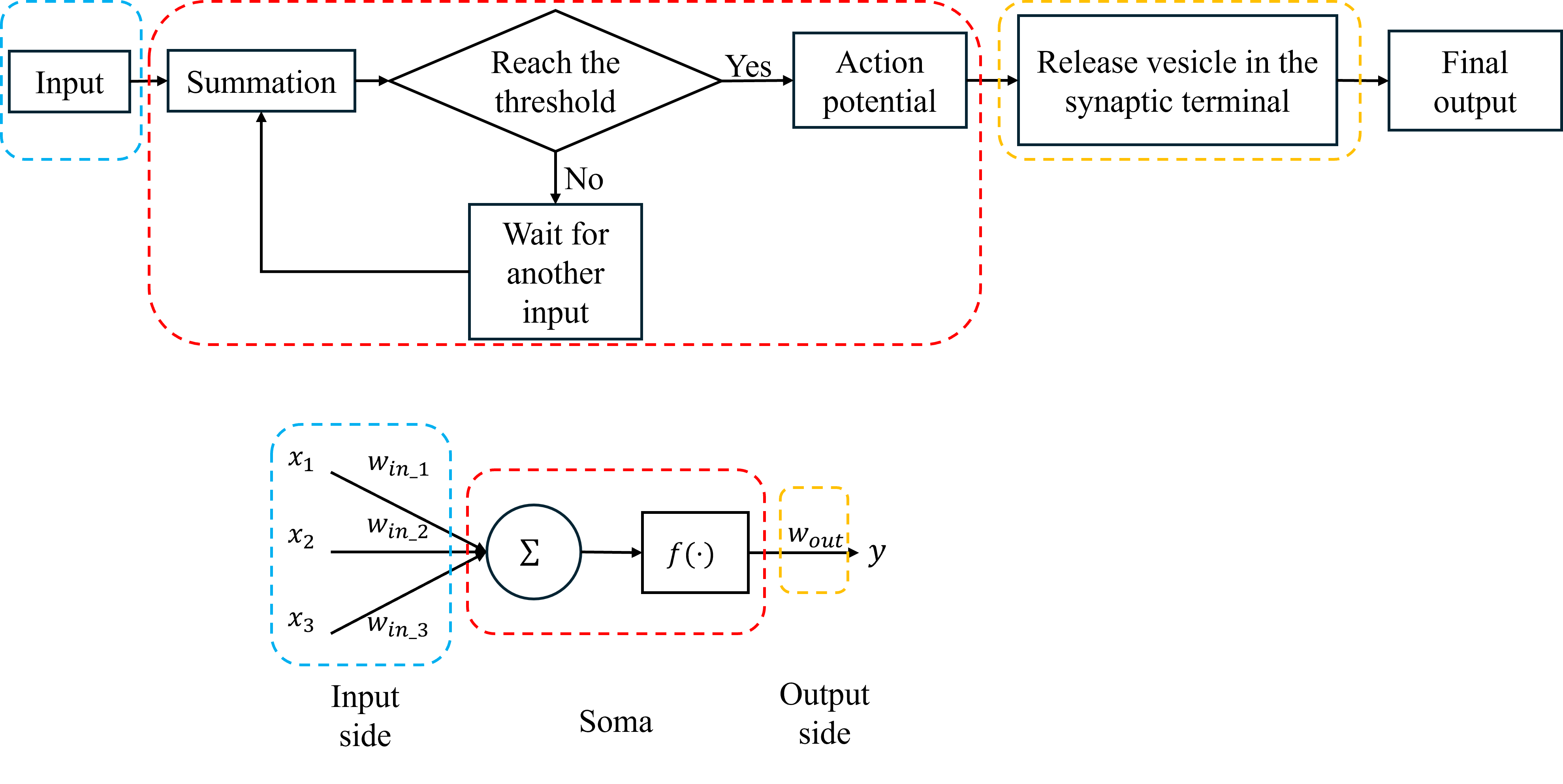}
    \caption{The upper panel illustrates the biological process abstracted by Purin, including input integration, threshold-triggered activation, and neurotransmitter release. The lower panel shows the corresponding computational abstraction of a single Purin neuron, which computes a weighted input sum, applies the signed magnitude function ($f(\cdot)$), and modulates the result with the output-side efficacy ($w_{out}$). The short-term factor ($g_{stp}$) is omitted because it represents a dynamic efficacy state rather than a distinct anatomical structure.}
    \label{fig:PurinNeuron}
\end{figure}

In this section, we propose a new abstraction for biological neurons that is compatible with the BP algorithm. We use a Purin neuron shown in Fig. \ref{fig:PurinNeuron} as an example to illustrate how we abstract the biological neuron into the proposed Purin mechanism, which illustrates the conceptual structure of a single Purin neuron, including its input side structure, soma, and output side structure. In a fully connected layer, multiple Purin neurons are implemented in the vector form, as described by Eq (\ref{general}). The Purin neuron uses a time interval abstraction (introduced in Section \ref{steptoperiod}), thereby simplifying the potential-based procedure in biological neurons into two steps that are close to conventional ANNs: 1) summarizing the weighted inputs from the input side and 2) applying a nonlinear function to the resulting weighted sum. For the illustrated Purin neuron, the result of $f(\cdot)$ is then multiplied by an output-side efficacy $w_{out}$ and a short-term factor $g_{stp}$, which is omitted from Fig. \ref{fig:PurinNeuron} because it represents a dynamic efficacy state rather than a distinct anatomical structure, as a simplified procedure for the release of vesicles in the synaptic terminal. Consequently, the output of the Purin neuron illustrated in Fig. \ref{fig:PurinNeuron} is calculated by Eq (\ref{Purinoutput}). Eq (\ref{ConvPurinneuron}) shows a Purin convolution layer with $C_{out}$ output channels. For the $j^{th}$ output channel, the $w_{out\_j}$ and $g_{stp\_j}$ are shared within each output channel. For notational consistency, we use the equivalent matrix multiplication form \cite{im2col} to express the convolution operation in our paper.
\begin{equation}
    \label{Purinoutput}
    y=w_{out}\times g_{stp}\times f(\sum^{3}_{i=1} w_{in\_i}x_i)
\end{equation}
\begin{equation}
    \label{ConvPurinneuron}
Y_j =w_{out\_j}\times g_{stp\_j}\times f(W_{in\_j}X), j=1,2,\ldots,C_{\mathrm{out}}
\end{equation}

\subsubsection{Output Efficacy}
We start by observing that signal transmission across a chemical synapse depends on the neurotransmitter, which is affected by the efficacy of the synaptic terminal. The proposed mechanism abstracts it as the output-side efficacy of ANN neurons, and introduces an efficacy factor $W_{out}$ to each neuron after the activation function. The proposed scheme does not take into account the decay of the signal during transmission because it is difficult to define the distance between neurons in ANNs. The equation of the classic ANN neuron is now rewritten as Eq (\ref{newEq}), where $f(\cdot)$ is the activation function. As $W_{out}$ represents the efficiency of transmitting the signal in presynaptic neurons, it should preserve the sign of the result of $f(\cdot)$. Therefore, in Purin, elements in $W_{out}$ are constrained to a non-negative number so that it scales the result of $f(\cdot)$ without reversing its sign.

\begin{equation}
    \label{newEq}
    Y=W_{out}\odot f(W_{in}X)
\end{equation}

\subsubsection{From Time Step to Time Interval}
\label{steptoperiod}
As most ANN models for image classification do not use temporal inputs and outputs, Purin requires an abstraction that introduces short-term dynamics without constructing discrete time steps. However, a single still image cannot form a temporal image sequence from the view of the ANN setting. On the other hand, research on rapid serial visual presentation suggests that biological visual recognition occurs over a short presentation interval rather than at an instantaneous moment \cite{RSVP1, RSVP2}. This conclusion provides the idea that an image can be abstracted as information presented in a short time interval.

The abstraction of input images is also required to introduce long-term and short-term plasticity to ANN models. In ANN architectures, the trainable weights remain unchanged within a training batch and are updated after one training batch is completed. This batch-based update can be treated as long-term synaptic efficacy changes in neural activities. But biological synapses can exhibit short-term efficacy changes before long-term synaptic changes are established. Therefore, if we want to introduce short-term synapse plasticity to ANNs while keeping BP-based long-term optimization, the feedforward process should be described using an interval-based abstraction rather than an instantaneous event.

Based on this view, we define the input images as the information presented in a short time interval. The time interval does not mean a specific physical duration. It represents a period in which long-term plasticity does not occur in the long-term weight matrices $W_{in}$ and $W_{out}$, while the short-term factor $g_{stp}$ changes according to the activity generated by the current input. After a training batch is completed, $W_{in}$ and $W_{out}$ update their value based on the gradient accumulated in the batch using the BP algorithm like ANNs, and $g_{stp}$ is reset to its default value. This abstraction allows Purin to introduce short-term and long-term plasticity mechanisms without spike encoding or simulating discrete time steps.

To meet the time interval abstraction, the activation function $f(\cdot)$ in ANN should not only represent the neural activation at a specific moment. In Purin, we interpret $f(\cdot)$ as a signed activity measure rather than as the absolute firing rate of a biological neuron. We name $f(\cdot)$ as signed magnitude function. Suppose a neuron has a baseline firing rate $M$ during a time interval. If its firing rate is higher than $M$ in a time interval, this neuron shows excitatory-like activity relative to its baseline. If its firing rate is lower than $M$ in a time interval, this neuron shows inhibitory-like activity. Since Purin does not model the firing rate, rather it abstracts the activity relative to the baseline, we can centre the baseline value $M$ at zero. If the value of $f(\cdot)$ is greater than 0, it indicates that the neuron's activity is above the baseline. If the value of $f(\cdot)$ is smaller than 0, it indicates that the neuron's activity is below the baseline. The absolute value of $f(\cdot)$ reflects the magnitude of the deviation. According to this interpretation, a neuron can have a negative or positive output in the proposed abstraction, which shows whether its activity is below or above its baseline. 

Since $f(\cdot)$ abstracts the neural activity in a short time interval, the weight matrices $W_{in}$, $W_{out}$ and the short-term factor $g_{stp}$ represent the efficacy during information transmission. In this light, the output of a Purin neuron can also be interpreted as transmitted activity over a short interval. Thus, we can apply this abstraction from the first layer to other deeper layers in the neural network.

\subsection{Short-term Plasticity}
The short-term plasticity is an important mechanism that refers to temporary changes in synaptic efficacy over short time scales. These changes can arise from presynaptic and/or postsynaptic mechanisms. For a biologically faithful model, we may need to rewrite Eq (\ref{newEq}) into Eq (\ref{strictstp}) shown below, where $\Delta w_{in\_i}$ and $\Delta w_{out\_i}$ are the efficacy changes caused by short-term plasticity.
\begin{equation}
    \label{strictstp}
     Y=(W_{out}+\Delta W_{out})\odot f((W_{in}+\Delta W_{in})X)
\end{equation}

However, this equation introduces a challenging problem that requires distinguishing where the short-term plasticity occurs and quantifying its strength. This issue is difficult because error signals in ANNs, such as gradients, are rarely applied to the formulation with split weights. Therefore, it is difficult to distinguish which component of the synaptic connection is affected by the error signals. To address this issue, we introduce a factor $g_{stp}$ to represent the efficacy changes caused by short-term plasticity and form the equation (\ref{Purinoutput}) for the output calculation in a Purin neuron.

Based on the time interval abstraction in \ref{steptoperiod}, $g_{stp}$ updates in the feedforward procedure for every input image. Eq (\ref{stpupdate}) shows how $g_{stp}$ updates, where $\lambda$ is the learning rate.
\begin{equation}
    \label{stpupdate}
    g_{stp\_new}=g_{stp\_old}+\lambda f(W_{in}X)
\end{equation}

To model the recovery of short-term efficacy changes, $g_{stp}$ is gradually moved towards its default value $d$. Eq (\ref{stpvanish}) shows how $g_{stp}$ recovers to the default value over time, where $d$ is the default value of $g_{stp}$ and $b$ is the batch size. The BP-based weight update (i.e., the long-term plasticity in Purin) triggers when a batch of samples is completed, therefore we abstract one training batch as a long time interval composed of $b$ short time intervals. 
\begin{equation}
    \label{stpvanish}
    g^{\prime}_{stp}=g_{stp}+(d-g_{stp})\times(1-exp({-1/b}))
\end{equation}

For 2D convolution layers, we average the sum of the function outputs (i.e., the sum of $f(\cdot)$ in a channel) to update $g_{stp}$ in a channel. For a 2D convolution neuron with $N$ elements in its output matrix, Eq (\ref{convstp}) shows how to update $g_{stp}$ in the feedforward process, where $s$ denotes the sum of the function outputs and $\lambda$ is the learning rate.
\begin{equation}
    \label{convstp}
     g_{stp\_new}=g_{stp\_old}+\lambda s/N
\end{equation}

In Purin, the short-term factor is first updated after processing each input using Eq (\ref{stpupdate}) or (\ref{convstp}), followed by the recovery operation shown in Eq (\ref{stpvanish}). To avoid $g_{stp}$ becoming infinitely large or small, we set a boundary for it, which is limited from 0.8 to 1.2 to provide a mild modulation factor for the output. Unlike test-time training or adaptation methods \cite{TTT}, the short-term factor $g_{stp}$ is updated by the activity-dependent rule and will not update the long-term trainable weights $W_{in}$ and $W_{out}$ during inference or validation.

\subsection{Recovery Analysis of Short-term Factor}
The dynamic short-term factor $g_{stp}$ affects the outputs of neurons in the Purin mechanism, and it is necessary to show its stability. Since we already set a boundary to $g_{stp}$, it does not lead the output $Y$ to infinity. Thus, in this section, we discuss whether it recovers to the default value $d$. We assume that the inputs of the neuron are 0 when analyzing the recovery of $g_{stp}$, which makes it clear whether the recovery step leads $g_{stp}$ to $d$.

Starting from Eq (\ref{stpvanish}) and denoting $1-exp(-1/b)$ as $\rho$, we need to measure the distance between $g^{\prime}_{stp}$ and the default value $d$. Thus, we subtract $d$ from both sides of Eq (\ref{stpvanish}) and rewrite it as Eq (\ref{stpminus}) by grouping $g_{stp}$ and $d$ on the right side.
\begin{equation}
    \label{stpminus}
    g^{\prime}_{stp}-d=(1-\rho) g_{stp}+(\rho-1)d
\end{equation}
It is obvious that $\rho-1=-(1-\rho)$, so Eq (\ref{stpminus}) can be rewritten as Eq (\ref{newstpminus}).
\begin{equation}
    \label{newstpminus}
    g^{\prime}_{stp}-d=(1-\rho)(g_{stp}-d)
\end{equation}
Since $\rho=1-exp({-1/b})$, where $b$ is the batch size (thus positive), $1-\rho=exp({-1/b})$ is in the range of $(0,1)$. Eventually, we can rewrite Eq (\ref{newstpminus}) to Eq (\ref{stpfinal}) that shows that the distance between $g^{\prime}_{stp}$ and the default value $d$ keeps shrinking after every recovery step.
\begin{equation}
    \label{stpfinal}
    g^{\prime}_{stp}-d=(g_{stp}-d)exp({-1/b})
\end{equation}

\subsection{Weight Update}
The weight matrices $W_{in}$ and $W_{out}$ can be trained by gradient descent as ANNs. Based on Eq (\ref{general}), we present how to update $W_{out}$ and $W_{in}$ in Eq (\ref{gwout}) to (\ref{win}), where $\eta$ is the learning rate.
\begin{equation}
    \label{gwout}
    \Delta W_{out}=\frac{\partial L}{\partial Y}\odot g_{stp}\odot f(W_{in}X)
\end{equation}
\begin{equation}
    \label{wout}
    W_{out}=W_{out}-\eta\Delta W_{out}
\end{equation}
\begin{equation}
    \label{gwin}
    \Delta W_{in}=\frac{\partial L}{\partial Y}\odot g_{stp}\odot W_{out} \odot f^{\prime}(W_{in}X) X^T
\end{equation}
\begin{equation}
    \label{win}
    W_{in}=W_{in}-\eta\Delta W_{in}
\end{equation}

In 2D convolution layers, the convolution kernel is shared within an output channel. As a result, there is one output weight for each kernel, and the equation for calculating $\Delta W_{out}$ is rewritten as Eq (\ref{cgwout}). In this equation, $Y$ is the output feature map for this channel, $\frac{\partial L}{\partial Y}$ is the backpropagation error signal, $N$ is the number of elements in $Y$, and $*$ denotes channel-wise multiplication.
\begin{equation}
    \label{cgwout}
    \Delta W_{out}=g_{stp}*\sum^N_{i=1} (f(W_{in}X) \odot \frac{\partial L}{\partial Y})
\end{equation}
The calculation of elements in $\Delta W_{in}$ is very similar to Eq (\ref{gwin}), the only difference is that $g_{stp}$ is shared within a channel.

\subsection{The Bias Vector and Dropout Mechanism}
In ANNs, the bias vector \pmb{$b$} inside the activation function provides flexibility to the neuron. From the perspective of biological abstraction, the bias vector plays a role similar to the firing threshold. However, this potential explanation of the bias vector works when simulating neuron's activities at a specific moment, which is incompatible with the time interval abstraction in the proposed mechanism. Although we attempted to identify a biologically plausible abstraction for the bias vector under our time period condition, we have not yet found a satisfactory answer. Therefore, we exclude the bias vector from our mechanism.

The dropout mechanism \cite{dropout} is a commonly used mechanism to prevent ANNs from overfitting by randomly setting the output of a neuron to 0 according to a probability $p$. Although this mechanism can be loosely related to unreliable neural transmission, it scales the weight during testing, which is a mathematical approximation for deterministic inference results rather than a biological mechanism. More importantly, $g_{stp}$ updates its value based on the inputs of the neuron. If the dropout mechanism is applied, the sum of the inputs will become stochastic due to the random mask, which leads $g_{stp}$ to be updated based on corrupted activity rather than the true activity from presynaptic neurons. Therefore, we do not preserve the dropout mechanism in Purin.

\section{Experiments}
\label{experiments}
In this section, we introduce our method to three fundamental deep learning models, AlexNet \cite{AlexNet}, VGG11 \cite{vgg}, and GoogLeNet \cite{googlenet} to evaluate Purin's performance. We replace the vanilla weight matrices with our split weights mentioned in Section \ref{method} to establish Purin version models.

\subsection{Datasets and Experimental Settings}
We evaluated the proposed mechanism on four widely used datasets: CIFAR-100 \cite{cifar100}, AID \cite{AID}, UCM \cite{UCM}, and 102 Category Flower Dataset \cite{Oxford102}. The relevant details of them are briefly listed below.

\begin{enumerate}
    \item CIFAR-100: this dataset contains 50000 training samples and 10000 test samples from 100 categories, with a size of $32\times32$ pixels in each image.
    \item AID: this dataset contains 10000 aerial images from 30 categories, with a size of $600\times600$ pixels in each image.
    \item UCM: this dataset contains 2100 aerial images from 21 categories, with a size of $256\times256$ pixels in each image.
    \item 102 Category Flower Dataset: This dataset  contains 8189 flower images with various image sizes. Each class consists of between 40 and 258 images. In this paper, we denote this dataset as Oxford102.
\end{enumerate}

We resized the images in all four datasets to $224\times224$ pixels to match the input shape of AlexNet and GoogLeNet architectures. We resized the images in the AID, UCM, and Oxford102 datasets to $224\times224$ pixels for VGG11 to avoid the out-of-memory problem on our device. We randomly selected 10\% of the training samples in the CIFAR-100 dataset as the validation set. The AID and UCM datasets are divided into training, validation, and test sets with proportions of 60\%, 20\%, and 20\%, respectively. We used the official split of the Oxford102 dataset which contains 1020 images in the training and validation set, respectively, and 6149 images in the test set.

All experiments were run on a computer with a NVIDIA RTX A5000 graphic card and 24 GB of memory. The coding environment for vanilla and matched models included Ubuntu 24.04, Python 3.11, and PyTorch 2.4.0 with CUDA 11.8. We implement our method on the same device in C/C++ and CUDA which was compiled with a NVCC compiler v11.7. The batch size was set to 100. The AdamW optimizer with a learning rate of 0.0001 and a weight decay of 0.0005 was used for all models in our experiments. We used Leaky ReLU ($\alpha=0.01$) as the signed magnitude function $f(\cdot)$ for the models that introduced our mechanism. The learning rate $\lambda$ for $g_{stp}$ was set to 0.0001. The default value $d$ for $g_{stp}$ was set to 1.0. Input images were scaled to the range $[-1.0, 1.0]$ for models with Purin mechanism. The elements in the output synapse matrices $W_{out}$ were initialized from a uniform distribution U(0.9, 1.1). The $g_{stp}$ and $W_{out}$ were disabled in the last fully connected layer, which meant that their values were always 1, to ensure stable outcomes. In Purin-based models, we did not use batch normalization. This design focuses on the effect of split weight and short-term factor in Purin without introducing additional normalization mechanisms. The short-term factor $g_{stp}$ was reset to its default value (1.0) after each batch during training, validation, and inference. It was also reset before each training epoch and before the validation and inference stages began. Therefore, $g_{stp}$ only retained the short-term state within a batch and did not carry information across batches or different stages.

Although the pre-trained weights of the aforementioned models could be loaded into models with our mechanism, they are obtained without the split weight mechanism. As a result, they will mismatch the distribution of input features under our mechanism due to the changes of the activation function and neuron's input. Thus, we decided to train all models from scratch to make fair comparisons.

\subsection{Analysis of Short-term Factor}
In this section, we analyse the behaviours of the short-term factor $g_{stp}$ proposed in this paper. We present experimental results on different $g_{stp}$ variants, $g_{Stp}$ strength among samples, and its changes under different input sequences. 
\subsubsection{Validation of Short-term Factor}
In this section, we compare four variants of $g_{stp}$, which are indicated as variants A to D in Table \ref{tab:gstpvariant}, with the one using the full Purin mechanism (variant E) on the CIFAR-100 and UCM datasets. The variant A means $g_{stp}$ is generated from U(0.8,1.2) after one sample is completed. We used the VGG11 and GoogLeNet architectures for the experiments. The experimental results over five independent experiments are listed in Table \ref{tab:gstpresult}.
\begin{table}[pos=h]
    \centering
    \caption{Variants of $g_{stp}$}
    \begin{tabular}{ccc}
    \toprule
        Variant &  exponential recovery& range\\
    \midrule
        A &- &$g_{stp}\sim U(0.8,1.2)$\\
        B&-&(0.8,1.2)\\
        C&\checkmark& $(0, +\infty)$\\
        D&-&$(0, +\infty)$\\
        E&\checkmark& $(0.8, 1.2)$\\
    \bottomrule
    \end{tabular}
    \label{tab:gstpvariant}
\end{table}

\begin{table}[pos=h]
    \centering
    \caption{OA for architectures with different $g_{stp}$ variants}
    \resizebox{0.8\linewidth}{!}{
    \begin{tabular}{cccc}
    \toprule
     Architecture&Variant& CIFAR-100&UCM \\
     \midrule
        VGG11 & A&$37.64\pm0.43\%$&$62.19\pm1.64\%$\\
        &B&$38.60\pm0.70\%$&$65.71\pm2.31\%$\\
        &C&$38.92\pm0.41\%$&$66.05\pm1.22\%$\\
        &D&$38.60\pm1.02\%$&$67.24\pm1.71\%$\\
        &E&\pmb{$39.81\pm0.84\%$}&\pmb{$68.23\pm2.61\%$}\\
        \midrule
        GoogLeNet& A&$30.50\pm0.30\%$&$58.19\pm1.77\%$\\
        &B&$52.80\pm0.93\%$&$73.62\pm4.65\%$\\
        &C&$52.06\pm0.91\%$&$73.33\pm2.25\%$\\
        &D&$51.93\pm1.02\%$&$73.57\pm1.09\%$\\
        &E&\pmb{$52.86\pm0.53\%$}&\pmb{$74.05\pm1.75\%$}\\
    \bottomrule
    \end{tabular}
    }
    \label{tab:gstpresult}
\end{table}
According to the results in Table \ref{tab:gstpresult}, variant A always achieves the lowest OA on the two datasets for both VGG11 and GoogLeNet architectures. Although the OA gap between variant A and other variants is only 2-6\% for the VGG11 architecture, it becomes much larger for GoogLeNet where variant A performs approximately 18\% worse than other variants. This result suggests that $g_{stp}$ is not equivalent to random noise and provides better results. The OA values of the other four variants are similar, with differences of approximately 1\%, but variant E achieves the highest accuracy. In addition, the performances of variants B, C, and D are mixed in these experiments, and it is difficult to tell under what conditions these variants are better choices. However, the mechanisms applied to $g_{stp}$ in Purin help models achieve consistent performance across different architectures and datasets. Therefore, exponential recovery and limited value range of $g_{stp}$ provide more stable and better performance in classification accuracy when combined with each other.

\begin{figure*}
    \centering
    \subfigure[]{
    \includegraphics[width=0.45\textwidth]{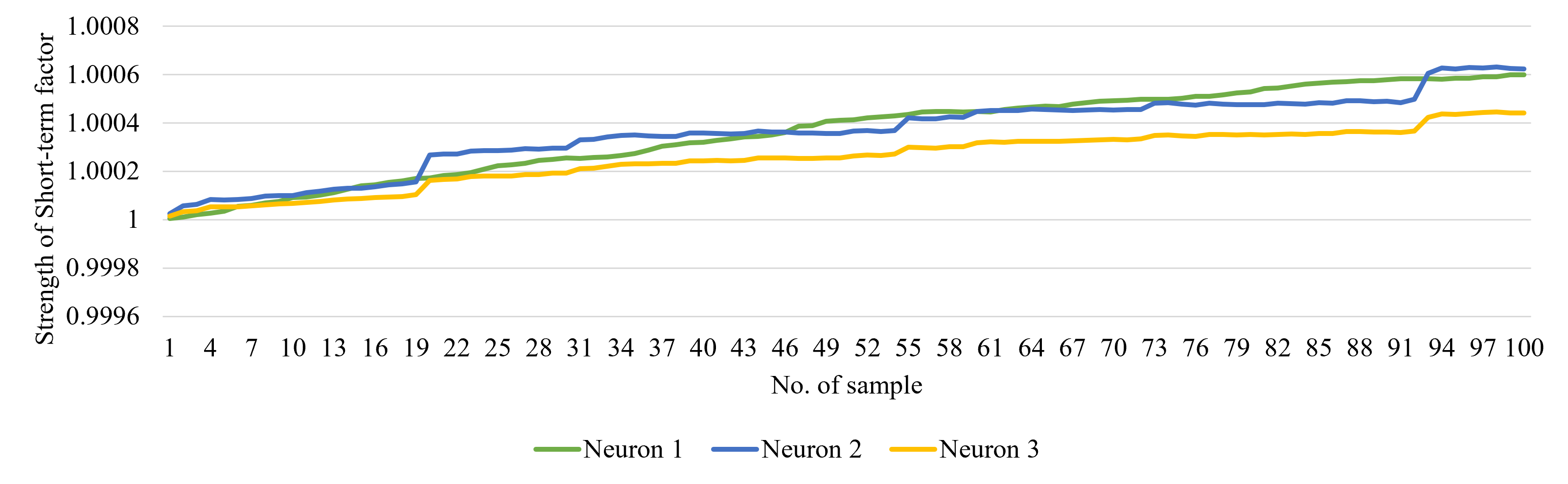}}
    \subfigure[]{
    \includegraphics[width=0.45\textwidth]{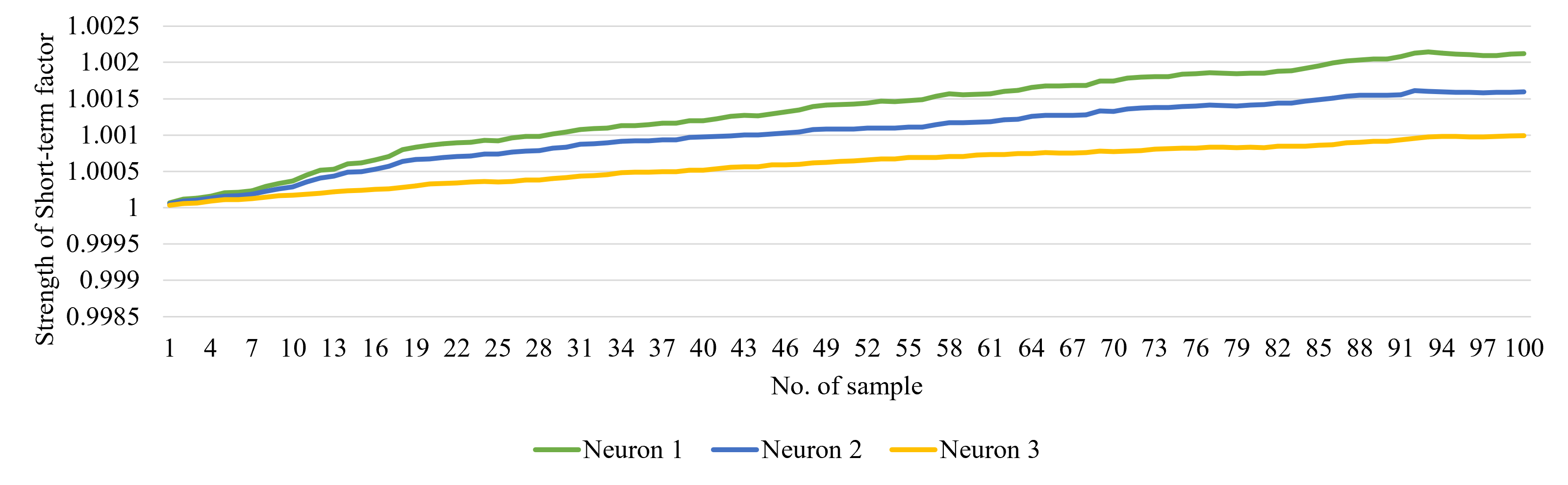}
    }
    \subfigure[]{
    \includegraphics[width=0.45\textwidth]{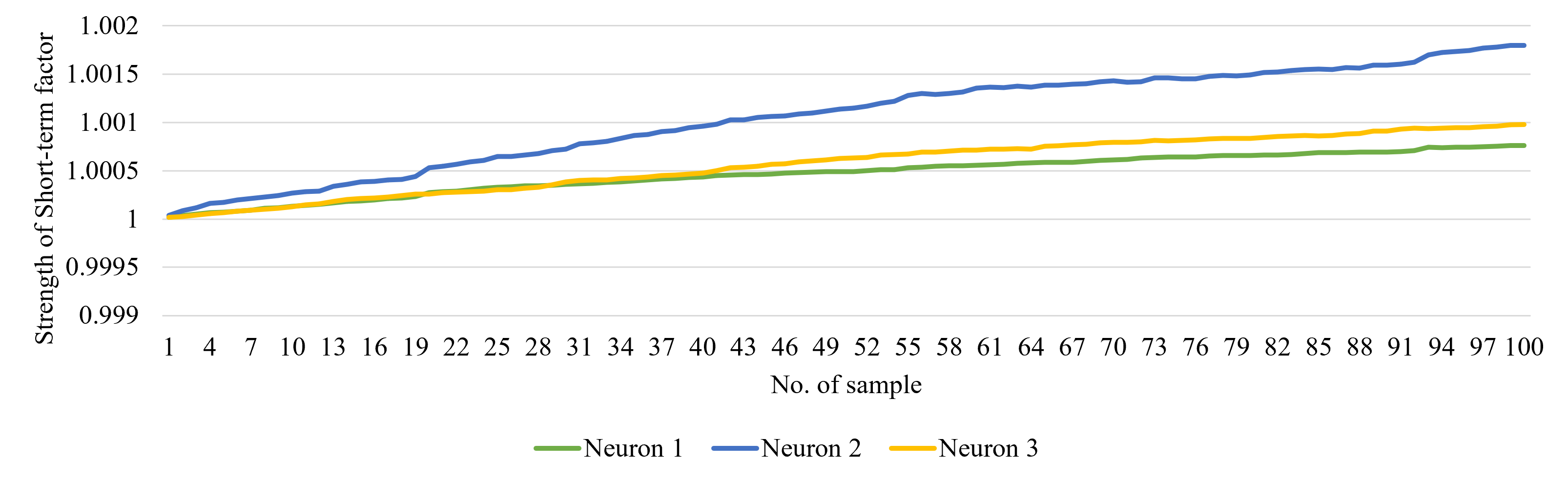}
    }
    \subfigure[]{
    \includegraphics[width=0.45\textwidth]{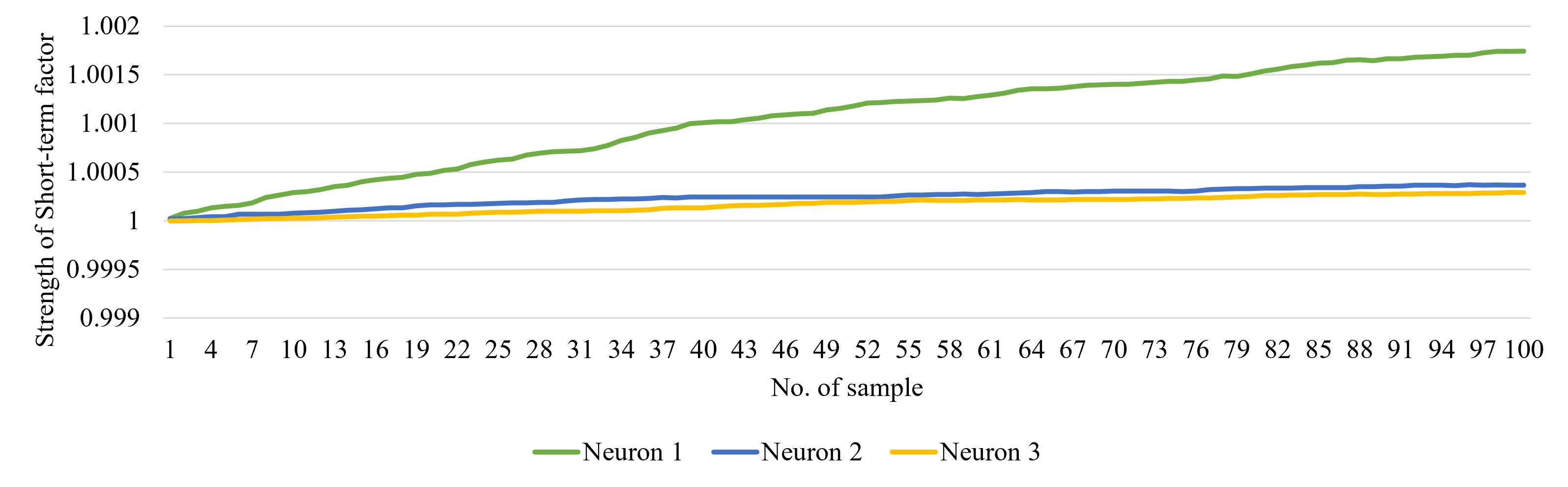}
    }
    \subfigure[]{
    \includegraphics[width=0.45\textwidth]{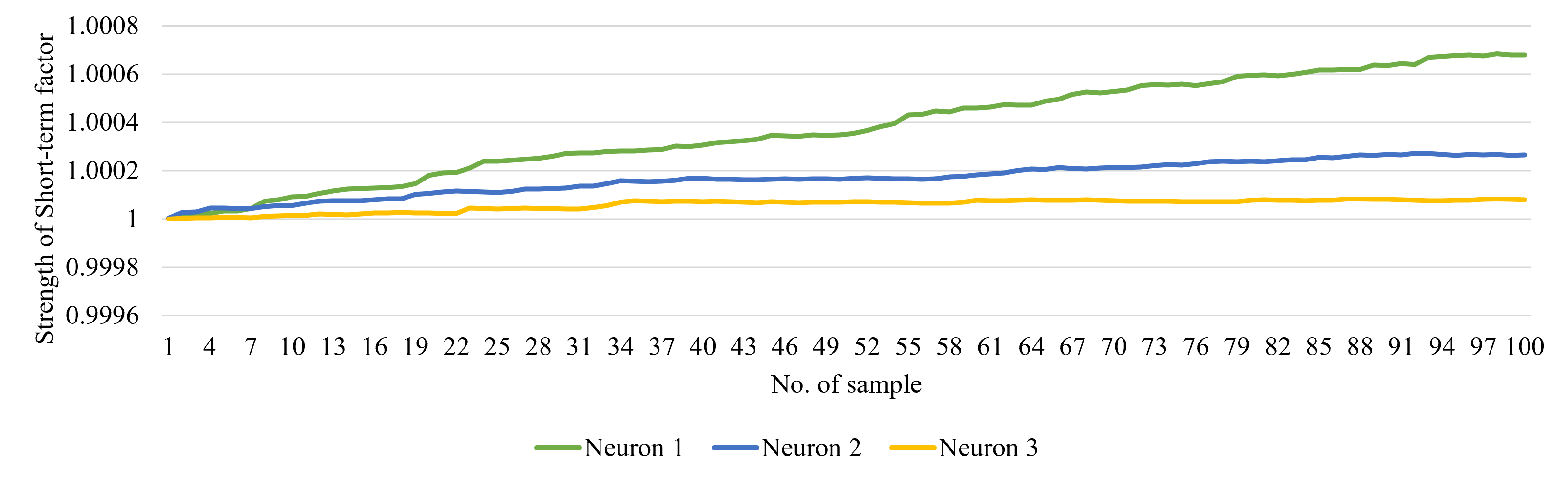}
    }
    \subfigure[]{
    \includegraphics[width=0.45\textwidth]{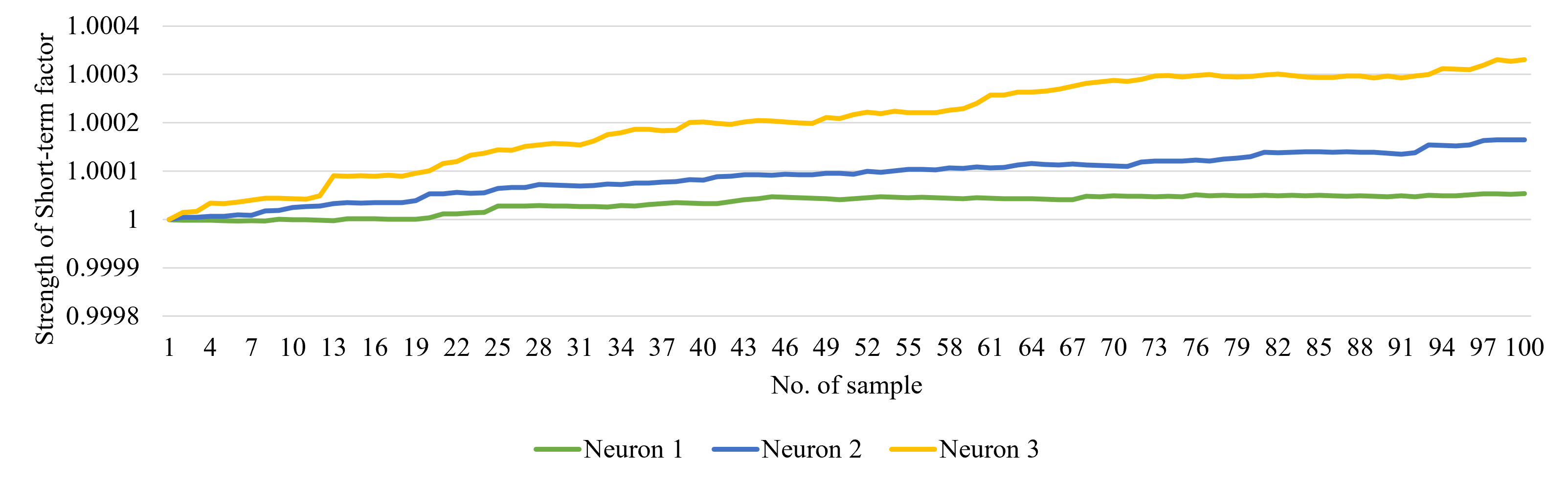}
    }
    \subfigure[]{
    \includegraphics[width=0.45\textwidth]{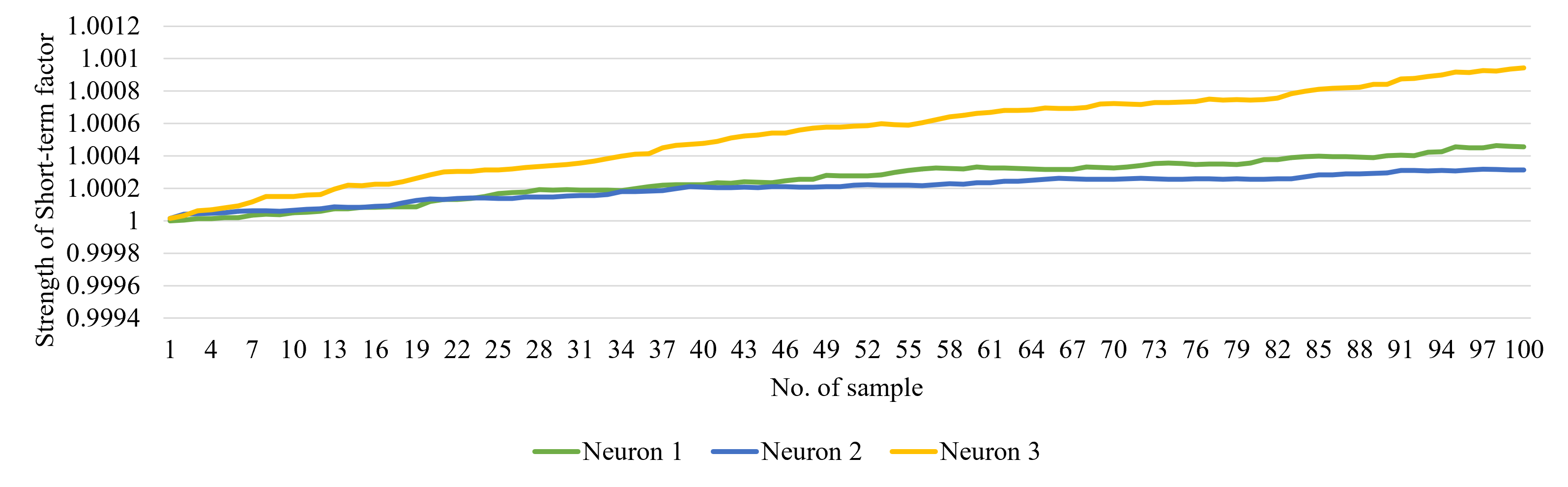}
    }
    \subfigure[]{
    \includegraphics[width=0.45\textwidth]{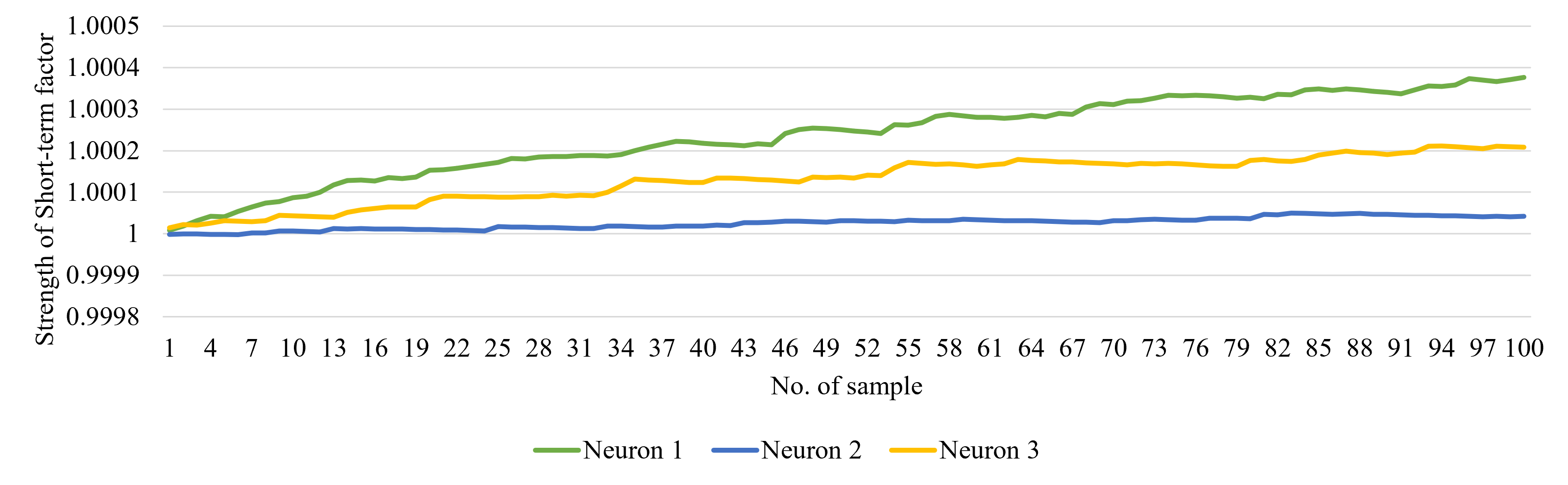}
    }
    \subfigure[]{
    \includegraphics[width=0.45\textwidth]{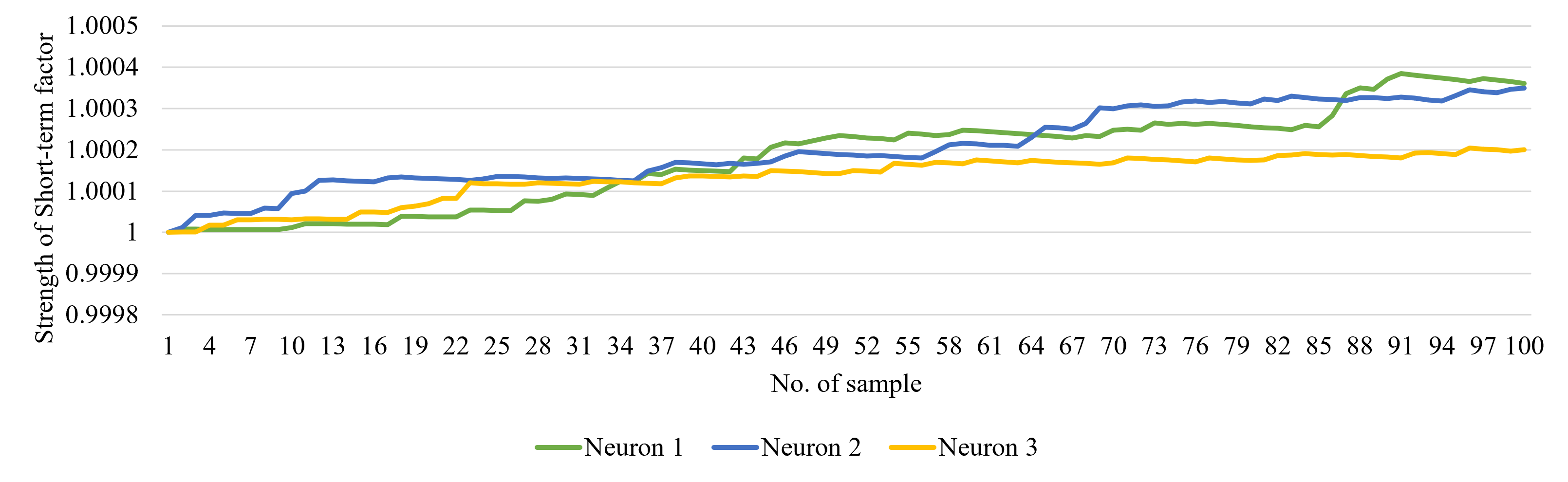}
    }
    \subfigure[]{
    \includegraphics[width=0.45\textwidth]{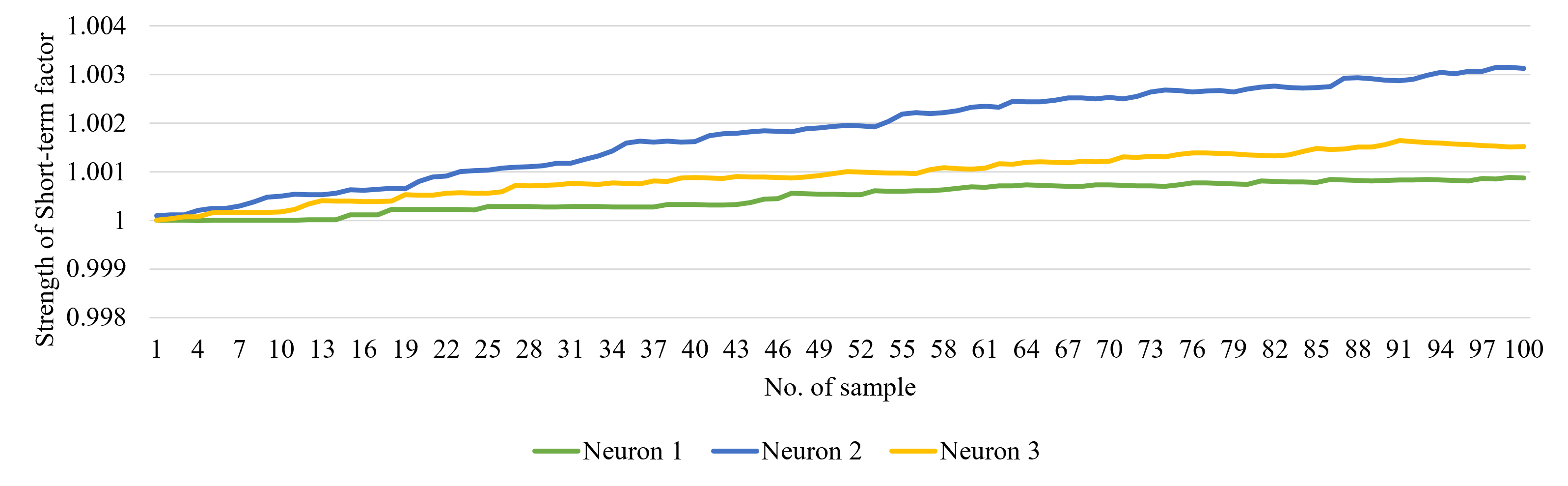}
    }

    \caption{$g_{stp}$ curves in an evaluation batch in convolution layers and fully connected layers in the Purin+VGG11 model. The last fully connected layer does not use $g_{stp}$. (a) Convolution\_0. (b) Convolution\_2. (c) Convolution\_4, (d) Convolution\_5. (e) Convolution\_7. (f) Convolution\_8. (g) Convolution\_10. (h) Convolution\_11. (i) fully connected layer\_14. (j) fully connected layer\_15.}
    \label{fig:gstpcurve}
\end{figure*}

We visualize the short-term factor $g_{stp}$ curve in a trained Purin+VGG11 model in an evaluation batch on the CIFAR-100 dataset \cite{cifar100}. We choose three channels/neurons from each convolution layer or fully connected layer as examples. We number the layers in VGG11 architecture from 0 to 16 (including all maxpooling layers and a flatten layer): Convolution\_0, Maxpooling\_1, Convolution\_2, etc. The $g_{stp}$ curves for each layer are shown in Fig. \ref{fig:gstpcurve}.

As shown in Fig. \ref{fig:gstpcurve}, the values of $g_{stp}$ in different layers do not saturate at either the upper or lower boundary within a batch. Therefore, the short-term factor provides mild scaling to the output of a neuron. The increasing trend of $g_{stp}$ can be attributed to the leaky ReLU function used in this paper, which scales the negative outputs by 0.01 and thus weakens their influence on $g_{stp}$ updates.

\subsubsection{History-dependent Behaviours of $g_{stp}$}
In this section, we estimate whether $g_{stp}$ introduces history-dependent computation. To achieve this goal, we constructed three input sequences with the same images but presented the input images in different orders and inferred the same target image after each sequence. We chose 100 images from the CIFAR-100 dataset to construct the input sequence and another image for inference. We froze $W_{in}$ and $W_{out}$ in all stages. Thus, order-dependent differences in the network arise from dynamic $g_{stp}$. We also repeated the experiments with disabled $g_{stp}$ (the values in $g_{stp}$ were always 1.0) as control experiments. In this section, we used the Purin + VGG11 architecture for experiments. Only for experiments in this section, we did not reset the value of $g_{stp}$ during inference. This setting was used to examine whether different input histories could produce different short-term states and lead to different outputs to the same target image.

\begin{table}[pos=h]
    \centering
    \caption{$L_2$ distances and maximum logit distances for outputs in the last layer}
    \begin{tabular}{cccc}
    \toprule
       Input order  & $g_{stp}$& $L_2^{(a,b)}$& $D_{max}^{(a,b)}$ \\
    \midrule
       1 vs. 2& Enable&$3.0922 \times 10^{-4}$&$7.4\times 10^{-5}$\\
       1 vs. 3&Enable&$7.0777\times 10^{-4}$&$1.79\times 10^{-4}$\\
       2 vs. 3&Enable&$5.8910\times 10^{-4}$&$1.72\times 10^{-4}$\\
       \midrule
       1 vs. 2& Disable&0&0\\
       1 vs. 3&Disable&0&0\\
       2 vs. 3&Disable&0&0\\
    \bottomrule
    \end{tabular}
    
    \label{tab:logitdistance}
\end{table}
To avoid the nonlinear normalization caused by the softmax function, which leads to very small numeric differences between different orders, we choose the pre-softmax output logit vectors for comparisons. We present comparisons of the $L_2$ distance and the maximum logit distance (defined in Eq (\ref{L2_dis}) and (\ref{max_log_dis}), where $z_c^a$ and $z_c^b$ are values in two logit vectors from the last layer, C is the number of categories) for outputs in the last layer between different input orders in Table \ref{tab:logitdistance}. When $g_{stp}$ was enabled, the network generated different outputs for the same test image under different input sequences. These distances are small because the value of $g_{stp}$ slightly changes within a batch and brings mild changes to the outputs of neurons. In contrast, when $g_{stp}$ was disabled, the network generated the same outputs for different input sequences. These results indicate that the short-term factor retains information about previous neural activity and introduces history-dependent responses to the architecture.
\begin{equation}
    \label{L2_dis}
    L_2^{(a,b)}=\sqrt{\sum^C_{c=1}(z_c^a-z_c^b)^2}
\end{equation}
\begin{equation}
D_{max}^{(a,b)}=\max_{1 \leq c \leq C}|z_c^a-z_c^b|
    \label{max_log_dis}
\end{equation}

\begin{figure}
\centering
    \subfigure[]{
    \includegraphics[width=0.95\linewidth]{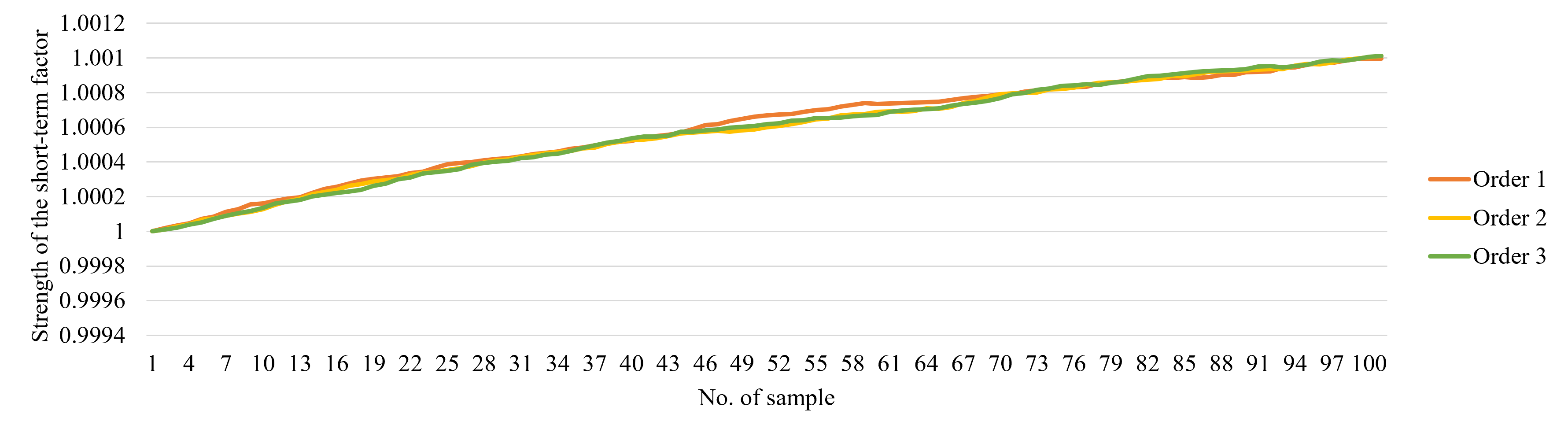}}
    \subfigure[]{
    \includegraphics[width=0.95\linewidth]{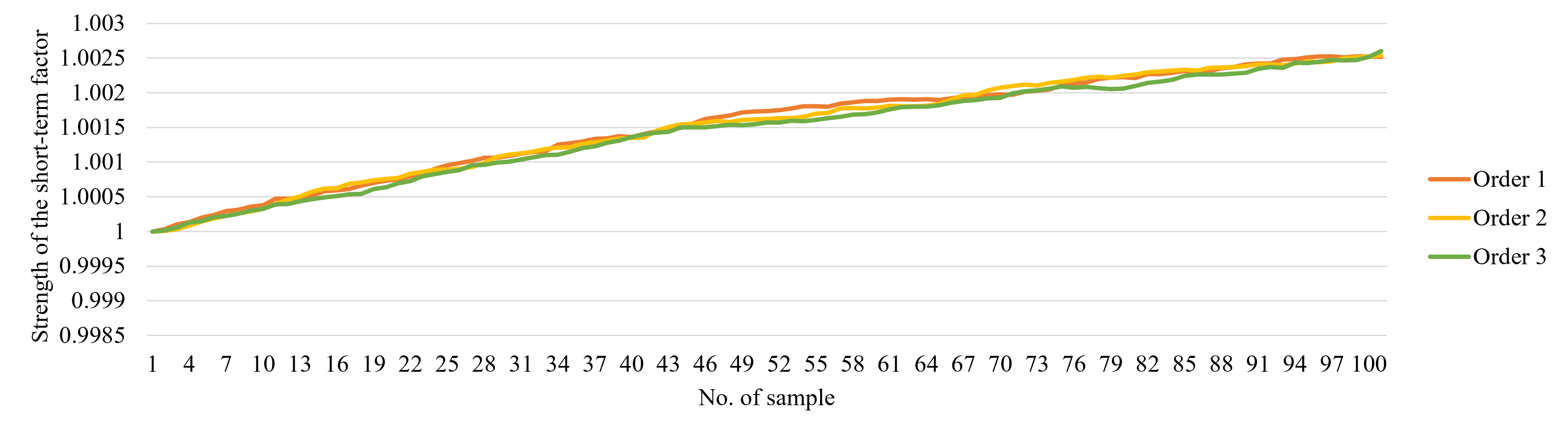}
    }
    \subfigure[]{
    \includegraphics[width=0.95\linewidth]{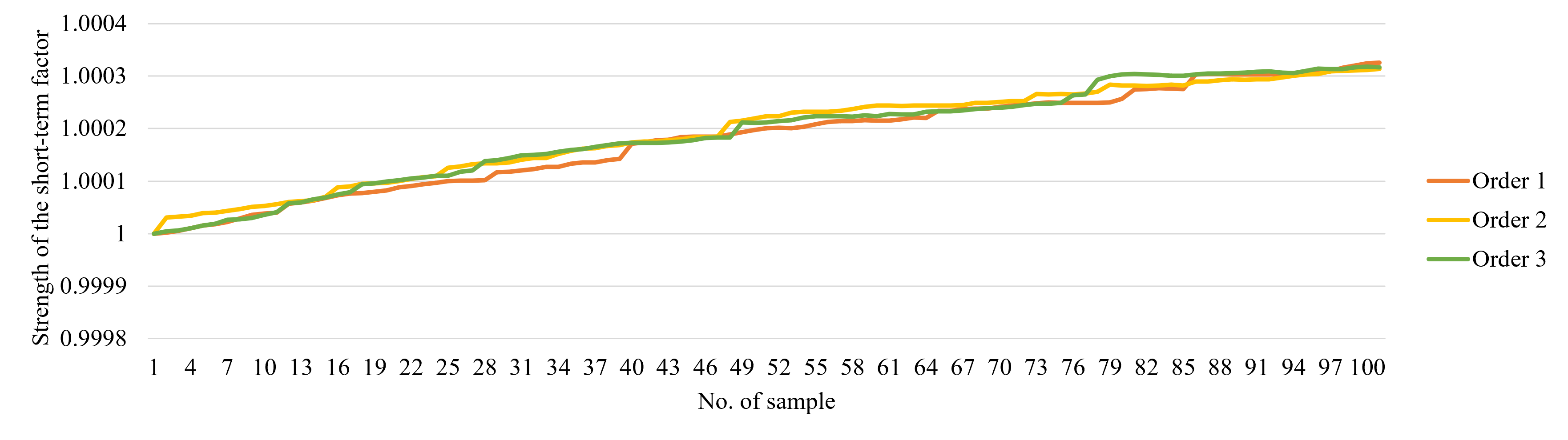}
    }
    \caption{$g_{stp}$ trajectories of three channels in the second convolution layer under different input order. (a) Channel 1. (b) Channel 2. (c) Channel 3.}
    \label{fig:gstp_3_neurons}
\end{figure}

Fig. \ref{fig:gstp_3_neurons} presents $g_{stp}$ curves of three Purin neurons in the second convolution layer under different input sequences. It is clear that the values of $g_{stp}$ are different between the input sequences, which means that these neurons received different inputs from the previous layer. These results also indicate that the differences in Table \ref{tab:logitdistance} originate from the history-dependent $g_{stp}$ states.

\subsection{Ablation Study}
In this section, we explore the effects of different parts in our mechanism. We denote different variants in Table \ref{tab:ablamodels} and list the experimental results over five independent experiments in Table \ref{tab:ablation}.

\begin{table}[pos=h]
\caption{Ablation variants}
    \centering
    \small
    \begin{tabular}{ccc}
    \toprule
         Variant& Split weight&$g_{stp}$\\
         \midrule
         A& -&- \\
         B&  \checkmark&-\\
         C&\checkmark&\checkmark\\
         \bottomrule
    \end{tabular}
    \label{tab:ablamodels}
\end{table}

\begin{table}[pos=h]
\caption{OA of ablation variants on four datasets}
    \centering
    \resizebox{\linewidth}{!}{
    \begin{tabular}{cccccc}
    \toprule
       Model&Variant  &  UCM&AID&CIFAR-100&Oxford102\\
       \midrule
AlexNet&A&$61.38\pm0.62$\%&$54.13\pm1.16$\%&$42.11\pm0.66$\%&$16.62\pm2.22$\%\\
       &B&\pmb{$72.43\pm0.94\%$}&$58.74\pm1.27\%$&$44.51\pm0.46\%$&$26.68\pm1.25\%$\\
       &C&$72.23\pm1.62\%$&\pmb{$59.42\pm1.05\%$}&\pmb{$45.94\pm0.41\%$}&\pmb{$26.83\pm0.90\%$}\\
       \midrule
VGG11&A&$51.33\pm3.44$\%&$50.27\pm1.37$\%&$33.31\pm1.13$\%&$10.55\pm0.77$\%\\
       &B&$65.42\pm2.53\%$&$56.73\pm0.85\%$&$38.09\pm0.93\%$&$23.50\pm0.67\%$\\
       &C&\pmb{$68.23\pm2.61\%$}&\pmb{$57.85\pm1.44\%$}&\pmb{$39.81\pm0.84\%$}&\pmb{$23.89\pm0.79\%$}\\
       \midrule
GoogLeNet&A&{$49.95\pm3.34$\%}&$53.80\pm1.57$\%&{$39.27\pm0.97$\%}&{$14.42\pm2.17$\%}\\
    &B&\pmb{$74.76\pm1.40\%$}&$61.73\pm0.99\%$&$47.74\pm0.70\%$&$27.00\pm1.91\%$\\
    &C&$74.05\pm1.75\%$&\pmb{$64.23\pm2.00\%$}&\pmb{$52.86\pm0.53\%$}&\pmb{$27.62\pm2.37\%$}\\
       \bottomrule
    \end{tabular}
    }
    \label{tab:ablation}
\end{table}

The experimental results indicate that the split weight mechanism contributes the most to the improvement of OA. Compared to variant A, in most cases, the classification accuracies of variant B for all three architectures are improved by at least 4\% on four datasets. The short-term factor $g_{stp}$ can further improve accuracy of variant B by approximately 0.2\% to 5\% in most cases, which is a smaller magnitude of improvement compared to that brought by the split weight mechanism.

In terms of architecture, the VGG11 architecture always benefits from the Purin mechanism on all four datasets, whereas the AlexNet and GoogLeNet architectures benefit from the proposed mechanism on all datasets except the UCM one. However, compared with variant B of the two architectures, the accuracy only decreases by 0.2\% for AlexNet and 0.71\% for GoogLeNet. These minor differences cannot provide strong evidence that the full Purin mechanism hurts the performance of the two architectures. Although the accuracies on the UCM slightly decrease, the $g_{stp}$ provides additional OA improvement over the split weight only variants on most datasets and architectures.


\subsection{Compatibility with Conventional ANN models}
\label{vp}
In this section, we investigate whether the proposed Purin mechanism is compatible with conventional ANN models. Thus, we compare the overall classification accuracy (OA) of vanilla models and models with our mechanism. The experimental results, which are the average accuracy over five independent experiments, are shown in Table \ref{tab:compare}. 

\begin{table}[pos=h]
    \centering
    \caption{OA on different datasets. The average accuracy and the standard deviation are given}
    \resizebox{\linewidth}{!}{
    \begin{tabular}{ccccc}
    \toprule
         Model&  UCM&AID&CIFAR-100&Oxford102\\
         \midrule
         AlexNet \cite{AlexNet}&$64.52\pm0.97\%$&\pmb{$61.33\pm1.84\%$}&\pmb{$48.92\pm0.56\%$}&$17.43\pm2.39\%$\\
         Purin+AlexNet&\pmb{$72.23\pm1.62\%$}&$59.42\pm1.05\%$&$45.94\pm0.41\%$&\pmb{$26.83\pm0.90\%$}\\
         
         VGG11 \cite{vgg}&$51.95\pm2.42\%$&$55.47\pm1.52\%$&$32.44\pm0.42\%$&$7.4\pm0.97\%$\\
         Purin+VGG11&\pmb{$68.23\pm2.61\%$}&\pmb{$57.85\pm1.44\%$}&\pmb{$39.81\pm0.84\%$}&\pmb{$23.89\pm0.79\%$}\\
         
         GoogLeNet \cite{googlenet}&\pmb{$76.76\pm2.57\%$}&$63.37\pm3.91\%$&\pmb{$55.05\pm0.67\%$}&\pmb{$33.25\pm3.20\%$}\\
         Purin+GoogLeNet&$74.05\pm1.75\%$&\pmb{$64.23\pm2.00\%$}&$52.86\pm0.53\%$&$27.62\pm2.37\%$\\
        \bottomrule
    \end{tabular}
    }
    \label{tab:compare}
\end{table}

The experimental results show that our mechanism improves the OA of VGG11 in all four datasets. Our mechanism significantly improves OA of VGG11 on the Oxford102 dataset approximately 3 times. It also improves OA of VGG11 by 16.28\% on the UCM dataset and 7.37\% on the CIFAR-100 dataset. When incorporated with AlexNet, our mechanism improves the OA on the UCM and Oxford102 datasets by 7.71\% and 9.4\%, respectively. However, on the AID and CIFAR-100 datasets, models with our mechanism show about 2\% to 3\% drop in OA. Results on the GoogLeNet architecture show that our mechanism cannot help improve OA on the UCM, CIFAR-100, and Oxford102 datasets, and only slightly improves the OA on the AID dataset by 0.86\%.

The experimental results shown above cannot be considered as fair comparisons because the Purin and vanilla models have differences in input pre-processing, activation function, and the use of the dropout and batch normalization mechanism (see Table \ref{tab:Mecdifference}). Therefore, it is difficult to identify whether the improvement and decrease in accuracy are caused by the proposed mechanism or not.
\begin{table}[pos=h]
    \centering
    \caption{Confounding factors between vanilla models and models with Purin}
    \resizebox{0.8\linewidth}{!}{
    \begin{tabular}{ccc}
    \toprule
         Confounding factor& Models with Purin& Vanilla models\\
         \midrule
       Input rescaling  & [-1.0, 1.0] & [0.0, 1.0]\\
       Activation function & Leaky ReLU & ReLU\\
       Use dropout& No& Yes\\
       Use batch normalization& No&Yes\\
       \bottomrule
    \end{tabular}
    }
    \label{tab:Mecdifference}
\end{table}

To estimate the effect of confounding factors, we evaluated matched models that use the same input rescaling, activation function, and dropout usage as models with Purin mechanism. The experimental results over five independent experiments are shown in Table \ref{tab:match}.
\begin{table}[pos=h]
    \centering
    \caption{Comparisons with matched models. The average accuracy and the standard deviation are given}
    \resizebox{\linewidth}{!}{
    \begin{tabular}{ccccc}
    \toprule
         Model & UCM&AID&CIFAR-100&Oxford102 \\
         \midrule
         AlexNet (matched)&$61.38\pm0.62$\%&$54.13\pm1.16$\%&$42.11\pm0.66$\%&$16.62\pm2.22$\%\\
         Purin+AlexNet&\pmb{$72.23\pm1.62\%$}&\pmb{$59.42\pm1.05\%$}&\pmb{$45.94\pm0.41\%$}&\pmb{$26.83\pm0.90\%$}\\

         VGG11 (matched)&$51.33\pm3.44$\%&$50.27\pm1.37$\%&$33.31\pm1.13$\%&$10.55\pm0.77$\%\\
         Purin+VGG11&\pmb{$68.23\pm2.61\%$}&\pmb{$57.85\pm1.44\%$}&\pmb{$39.81\pm0.84\%$}&\pmb{$23.89\pm0.79\%$}\\

         GoogLeNet (matched)&{$49.95\pm3.34$\%}&$53.80\pm1.57$\%&{$39.27\pm0.97$\%}&{$14.42\pm2.17$\%}\\
         Purin+GoogLeNet&\pmb{$74.05\pm1.75\%$}&\pmb{$64.23\pm2.00\%$}&\pmb{$52.86\pm0.53\%$}&\pmb{$27.62\pm2.37\%$}\\
         \bottomrule
    \end{tabular}
    }
    \label{tab:match}
\end{table}

Compared with matched models, Purin can improve OA for the matched models on all four datasets. For the Alexnet and VGG11 architectures, Purin improves the OA on the Oxford102 and UCM datasets by approximately 10\% for both models. On the AID and CIFAR-100 datasets, Purin improves the OA of the two models by approximately 3\% to 7\%. Without the dropout and batch normalization mechanisms, the OAs of the matched GoogLeNet architecture on four datasets significantly drop by 10\% to 26\%. With the help of the proposed mechanism, the accuracies of the Purin+GoogLeNet model reach a much closer value to the vanilla GoogLeNet (see Table \ref{tab:compare}), especially on the UCM and CIFAR-100 datasets where Purin helps recover the OA by 24.1\% and 13.59\%, respectively. Although the OA of the Purin+GoogLeNet model is still 5.63\% lower than the vanilla GoogLeNet on the Oxford102 dataset, it recovers 13.2\% of OA compared to the matched model on the same dataset.

\subsection{Additional Trainable Parameters and FLOPs Overhead Analysis}
We report the additional trainable parameter overhead of architectures with the Purin mechanism to ensure whether the increasing number of trainable parameters is the main reason for the improvement in OA. Suppose a convolutional layer has $C_{in}$ input channels, $C_{out}$ output channels, and a convolution kernel of size $k\times k$. The number of trainable parameters in this layer is $C_{out}C_{in}k^2$. For fully connected layers, the number of trainable parameters is $N_{in}\times N_{out}$, where $N_{in}$ and $N_{out}$ are the number of input neurons and output neurons, respectively. In Purin, we add an extra trainable weight $W_{out}$ to both the convolutional layers and the fully connected layers, which adds $C_{out}$ additional parameters to the convolutional layers and $N_{out}$ additional parameters to the fully connected layers. Therefore, the overhead of additional trainable parameters for each convolutional layer and fully connected layer in architectures with the Purin mechanism is calculated by Eq (\ref{convpara}) and Eq (\ref{fcpara}), respectively.
\begin{equation}
    \label{convpara}
    \frac{C_{out}}{C_{out}C_{in}k^2}=\frac{1}{C_{in}k^2}
\end{equation}
\begin{equation}
    \label{fcpara}
    \frac{N_{out}}{N_{in}N_{out}}=\frac{1}{N_{in}}
\end{equation}

The additional trainable parameter overheads for all three matched models used in this paper are presented in Table \ref{tab:overhead}. We assume that the size of the input image for all models is $224\times224$ and the number of categories is 100. Since the last fully connected layer in all three models disables $W_{out}$ and $g_{stp}$, it does not contribute to the number of additional parameters in the models.
\begin{table}[pos=h]
    \centering
    \caption{Relative overhead of additional trainable parameters for a dataset with $224\times224$ images, 100 categories}
    \begin{tabular}{ccc}
    \toprule
         Model& Extra trainable parameter & Overhead\\
         \midrule
         AlexNet&9344&0.0162\%\\
         VGG11&10944&0.0084\%\\
         GoogLeNet&7280&0.1199\%\\
         \bottomrule
    \end{tabular}
    
    \label{tab:overhead}
\end{table}

The additional trainable parameter overheads of Purin are less than 0.2\% compared to all three matched architectures. Therefore, the accuracy improvements in models with the Purin mechanism are unlikely to be explained simply by a large increase in the trainable model capacity.

In terms of FLOPs, the extra FLOPs in Purin come from multiplying $W_{out}$ and $g_{stp}$ during the feedforward process and the recovery and updating of $g_{stp}$. The additional FLOPs overheads in Purin models during inference are reported in Table \ref{tab:flops}, which are calculated using the same input resolution and number of categories as described above. We count one floating-point multiplication and one floating-point addition as one FLOP each. Therefore, one multiply-accumulate operation is counted as two FLOPs. Since the batch size is fixed and $1-exp(-1/b)$ in Eq (\ref{stpvanish}) is precomputed, we do not count this part in the FLOPs calculation. Compared with matched models, Purin introduces only a minor increase in inference FLOPs. Thus, the observed OA improvements are unlikely to be explained simply by the additional computational complexity.
\begin{table}[pos=h]
    \centering
    \caption{The overhead of FLOPs for a dataset with $224\times224$ images, 100 categories}
    \resizebox{\linewidth}{!}
    {
    \begin{tabular}{cccc}
    \toprule
         Model& Matched model FLOPs& Purin extra FLOPs& Increase \\
    \midrule
         AlexNet&1.421004 GFLOPs&1.518 MFLOPs&0.1068\%\\
         VGG11&15.210807 GFLOPs&22.349 MFLOPs&0.1469\%\\
         GoogLeNet&2.994909 GFLOPs&9.715 MFLOPs&0.3244\%\\
    \bottomrule
    \end{tabular}
    }
    \label{tab:flops}
\end{table}

\subsection{Comparisons with the SE block}
In \cite{SEblock}, authors developed the SE block that performs channel-wise multiplication on the output of the Inception module. This looks similar to the $g_{stp}\odot W_{out}$ operation in convolution layers in the GoogLeNet+Purin model. Thus, we evaluated the performance of the matched GoogLeNet architecture with SE blocks inserted after the convolution layer or inserted after the inception module as the original paper \cite{SEblock}. Since the GoogLeNet has only one fully connected layer as the classifier head, whose $W_{out}$ and $g_{stp}$ are disabled when incorporating with Purin mechanism, we can focus on the effect of the SE blocks and the $g_{stp}\odot W_{out}$ operation on the classification accuracy. The average OA and standard deviation over five independent experiments are shown in Table \ref{tab:SEvsgstp}. We also show the results of the matched GoogLeNet.
\begin{table}[pos=h]
    \centering
    \caption{OA on CIFAR-100 and UCM for different combinations of matched GoogLeNet}
    \resizebox{\linewidth}{!}{
    \begin{tabular}{ccc}
    \toprule
        Model & CIFAR-100&UCM \\
        \midrule
        Matched GoogLeNet+conv SE block & $38.43\pm1.16\%$&$56.43\pm0.97\%$\\
        Matched GoogLeNet+SE block&$35.96\pm1.97\%$&$57.71\pm4.32\%$\\
        Matched GoogLeNet&$39.27\pm0.97\%$&$49.95\pm3.34\%$\\       GoogLeNet+Purin&\pmb{$52.86\pm0.53\%$}&\pmb{$74.05\pm1.75\%$}\\
        \bottomrule
    \end{tabular}
    }
    \label{tab:SEvsgstp}
\end{table}

According to the results shown above, all three matched GoogLeNet models (with or without the SE block) do not outperform the combinations of GoogLeNet+Purin. Their OA are approximately 13\% to 25\% lower than that of the GoogLeNet+Purin variant on two datasets. The form of inserting SE blocks after all convolution layers is similar to how Purin applies $g_{stp}$ and $W_{out}$, but the OA of that combination is about 17\% to 20\% lower than that of GoogLeNet with full Purin mechanism on both datasets. These results indicate that the OA gains of Purin mechanism cannot be reproduced by simply inserting SE blocks into the matched GoogLeNet model.

\section{Conclusions and Future Work}
\label{conclusion}
We propose a biology-inspired mechanism, Purin, that introduces the output-side efficacy and the short-term plasticity to deep learning architectures for image classification tasks. The experimental results related to the short-term factor behaviours show that $g_{stp}$ provides mild modulation to the outputs of Purin neurons, which depends on the order of inputs. The short-term factor with  value range limitation and exponential recovery is more suitable for Purin neurons.

The ablation experiments indicate that, in most cases, the full Purin mechanism can further improve the classification accuracy compared to models that only use the split weight mechanism. Comparisons with the SE blocks suggest that the OA gains of Purin mechanism cannot be reproduced by simply inserting SE blocks into the matched GoogLeNet model.

The experimental results on compatibility show that even without transforming images to spike encoding like SNNs, Purin is compatible with the BP algorithm in ANNs due to its time interval based abstraction. After removing confounding factors from vanilla architectures, Purin models mechanism achieved higher classification accuracy on all four datasets for all matched models without significantly increasing the model size or the computational complexity.

This paper does not evaluate Purin with residual architectures such as ResNet18 \cite{resnet} due to differences in the neural structure. The extension of Purin to the residual connection mechanism is a possible research topic in future works.
\bibliographystyle{elsarticle-num}

\bibliography{Reference_Neurocomputing}





\end{document}